\documentclass[a4paper,fleqn]{cas-dc}

\usepackage[numbers]{natbib}
\usepackage{diagbox}
\usepackage{multirow}

\usepackage{graphicx}
\usepackage{subcaption}
\usepackage[utf8]{inputenc}
\DeclareUnicodeCharacter{202F}{\,}  

\def\tsc#1{\csdef{#1}{\textsc{\lowercase{#1}}\xspace}}
\tsc{WGM}
\tsc{QE}
\tsc{EP}
\tsc{PMS}
\tsc{BEC}
\tsc{DE}

\begin{document}
\let\WriteBookmarks\relax
\def\floatpagepagefraction{1}
\def\textpagefraction{.001}
\shorttitle{}
\shortauthors{Tao Wu et~al.}

\title [mode = title]{Understanding the Robustness of Graph Neural Networks against Adversarial Attacks}    

\author[1,2]{Tao Wu}[auid=000, orcid=0000-0003-1751-3040]
\affiliation[1]{organization={School of Computer Science and Technology, Chongqing University of Posts and Telecommunications}, 
    city={Chongqing},
    postcode={400065}, 
    country={China}}

\author[1]{Canyixing Cui}

\author[2]{Xingping Xian}
\cormark[1]
\ead{xxp0213@gmail.com}
\affiliation[2]{organization={School of Cyber Security and Information Law, Chongqing University of Posts and Telecommunications},
	city={Chongqing},
	postcode={400065}, 
	country={China}}
\cortext[cor1]{Corresponding author}

\author[3]{Shaojie Qiao}
\affiliation[3]{organization={School of Software Engineering, Chengdu University of Information Technology},
    city={Chengdu},
    postcode={610225}, 
    country={China}}

\author[4]{Chao Wang}
\affiliation[4]{organization={School of Computer and Information Science, Chongqing Normal University},
	city={Chongqing},
	postcode={401331}, 
	country={China}}

\author[2]{Lin Yuan}

\author[5]{Shui Yu}
\affiliation[5]{organization={School of Computer Science, University of Technology Sydney},
	city={Sydney},
	postcode={2007}, 
	country={Australia}}

\begin{abstract}
Recent studies have shown that graph neural networks (GNNs) are vulnerable to adversarial attacks, posing significant challenges to their deployment in safety-critical scenarios. This vulnerability has spurred a growing focus on designing robust GNNs. Despite this interest, current advancements have predominantly relied on empirical trial and error, resulting in a limited understanding of the robustness of GNNs against adversarial attacks. To address this issue, we conduct the first large-scale systematic study on the adversarial robustness of GNNs by considering the patterns of input graphs, the architecture of GNNs, and their model capacity, along with discussions on sensitive neurons and adversarial transferability. This work proposes a comprehensive empirical framework for analyzing the adversarial robustness of GNNs. To support the analysis of adversarial robustness in GNNs, we introduce two evaluation metrics: the confidence-based decision surface and the accuracy-based adversarial transferability rate. Through experimental analysis, we derive 11 actionable guidelines for designing robust GNNs, enabling model developers to gain deeper insights. The code of this study is available at \url{https://github.com/star4455/GraphRE}.
\end{abstract}


\maketitle
\section{Introduction}
\label{sec1}
Graph neural networks (GNNs), as a type of deep learning model designed to operate directly on graphs, have achieved remarkable success in various graph-based learning tasks. Since the graph convolutional network (GCN) was proposed in \cite{kipf2016semi}, many variants of GNNs have been developed, e.g., Convolutional-GNNs \cite{defferrard2016convolutional,wu2019simplifying}, Attention-GNNs \cite{velickovic2017graph}, Diffusion-GNNs \cite{gasteiger2018predict,liu2021graph}, Sampling-GNNs \cite{hamilton2017inductive}, and Isomorphism-GNNs \cite{xu2018powerful}. Despite their excellent performance, GNNs inherit some of the limitations of deep learning models, particularly their vulnerability to adversarial attacks. In particular, attackers craft adversarial examples by adding small perturbations to legitimate input samples, significantly degrading the performance of GNNs and inducing incorrect predictions, posing a substantial threat to their adversarial robustness in safety-critical applications. For instance, in cyberspace defense systems, GNNs have been widely utilized to detect a variety of attack behaviors. Adversarial attacks specifically designed to target GNNs can potentially undermine these defense systems, rendering them incapable of accurately identifying malware or detecting network intrusions \cite{caville2022anomal}. Similarly, in online social media platforms, there are a large number of fake accounts and fake news. Adversarial attacks on GNN-based detectors for fake accounts and fake news can undermine cybersecurity management, leading to the emergence of organized online fraud activities, unauthorized collection of information, and even interference in elections through the use of misleading content \cite{li2022sybilflyover}. Therefore, improving the robustness of GNNs against adversarial attacks can help prevent intelligent systems from generating erroneous or inaccurate prediction results, thereby enhancing their overall reliability and accuracy.

Currently, studies on the design of robust GNNs are increasingly comprehensive and cover various perspectives \cite{zhu2019robust, tang2020transferring, zhang2020gnnguard, shan2024gl}. However, existing studies are mainly based on empirical intuition and experimental trial and error, and it is unclear where adversarial robustness precisely comes from. Meanwhile, in contrast to the well-established and relatively mature framework against adversarial attacks on deep neural networks (DNNs) \cite{zhang2021detecting}, systematic research and a comprehensive understanding of the robustness of GNNs against adversarial attacks are still lacking. Due to the unique complexities introduced by graph-structured data, transferring insights from adversarial attacks on DNNs to GNNs is a nontrivial problem. Consequently, the guiding principles for enhancing the adversarial robustness of GNNs are currently missing. In order to provide principled guidance and design robust GNNs, there is an urgent need to conduct systematic research and gain an in-depth understanding of the adversarial robustness of GNNs.

With a given learning mechanism, the training process of deep learning models is influenced by training data, model architecture, and model capacity. However, how these factors affect the adversarial robustness of GNNs remains unclear, and there are still numerous challenges:  
(1) \textit{Limited understanding of graph patterns and their influence on the adversarial robustness of GNNs.} Training data is fundamental to learning GNNs, and the quality and quantity of this data directly impact the model's ability to make accurate predictions. For instance, intentionally introducing malicious perturbations to the features of a small number of nodes or to their connections could influence the representations generated by GNNs. Moreover, employing adversarial training techniques, which involve integrating adversarial examples into the training data, can effectively enhance the adversarial robustness of GNNs. However, how specific patterns in training data affect GNNs has rarely been systematically studied. This gap restricts our ability to fully leverage graph data for constructing robust GNNs. (2) \textit{Incomprehensive understanding of the role of model architecture in adversarial robustness.} The architecture of GNNs directly determines their expressive power and affects their representation results. For example, the aggregation mechanism iteratively updates the representation of each node by aggregating information from its neighboring nodes. Different aggregation mechanisms can produce different representations for the same data, which in turn affects the model's prediction results. However, there are few investigations into how different GNN variants learn graph representations and the manner by which the architecture of GNNs affects adversarial robustness is still ambiguous. 
(3) \textit{Insufficient insight into the impact of model capacity on adversarial robustness.} Model capacity refers to the amount of information that GNNs can learn and represent, impacting their ability to solve complex graph-related tasks. It is a crucial factor in determining how well GNNs can learn complex patterns and relationships within graph data. GNNs with insufficient capacity might struggle to learn intricate structures, while those with excessive capacity could overfit the training data. Meanwhile, adversarial attacks on graph data exploit the complex patterns and relationships within graphs to mislead GNNs. Thus, understanding how model capacity impacts the learning capabilities of GNNs is crucial for designing robust models. 

\textbf{Contribution.} Considering these challenges, we present the first comprehensive investigation of factors, i.e., graph patterns, model architecture, and model capacity, that may influence the adversarial robustness of GNNs. In particular, to explore the effects of graph patterns, we generate synthetic graphs with different structural regularities to train GNNs and analyze their adversarial robustness. We further compare these settings with an adversarial training strategy to demonstrate the effectiveness of diversifying data patterns by augmenting the training data with adversarial examples. Moreover, we analyze the adversarial robustness of different GNNs to identify which architecture contributes most to improving robustness against adversarial attacks. We also conduct a detailed analysis of neuron sensitivity across model architectures to gain deeper insights into the underlying mechanisms of adversarial attacks. Additionally, we investigate the adversarial robustness of GNNs with different capacities. Given that larger model capacity typically leads to more complex decision boundaries, we study the adversarial robustness of GNNs by simultaneously increasing model capacity and training data volume. We also measure the transferability of adversarial examples from the model perspective to understand how adversarial examples generated by one model influence other models. 

\begin{itemize}	
	\item \textbf{Comprehensive Adversarial Robustness Analysis Framework.} We propose a comprehensive empirical framework for thoroughly analyzing the adversarial robustness of GNNs (Fig.~\ref{fig1}). To the best of our knowledge, this is the first work to systematically investigate and compare the effects of graph patterns, model architecture, and model capacity on the adversarial robustness of GNNs. This systematic analysis uncovers key insights into the robustness behavior of GNNs and informs design principles that enhance resilience against adversarial attacks.
	
	\item \textbf{Novel Evaluation Metrics for Adversarial Robustness Analysis.} We propose two evaluation metrics to facilitate the analysis of adversarial behavior in GNNs: the confidence-based decision surface ${\mathcal S}\left( {\mathcal G} \right)$, which captures the model's sensitivity to structured perturbations and reflects its adversarial robustness; and the accuracy-based adversarial transferability rate (ATR), which quantifies the transferability of adversarial examples across different GNNs. 
	
	\item \textbf{11 Actionable Design Guidelines for Robust GNNs.} We conduct a systematic and comprehensive investigation into the adversarial robustness of GNNs across four benchmark datasets and three representative attack methods, leading to the distillation of eleven generalizable and actionable guidelines for designing robust GNNs. These principles reveal how graph patterns, model architecture, and model capacity influence adversarial robustness. 
\end{itemize}

The remainder of this paper is organized as follows. In Section 2, we review and summarize the related work. In Section 3, we introduce the problem definition and preliminary concepts. Section 4 presents a unified framework for exploring the adversarial robustness of GNNs, and Section 5 details the experimental analysis on robustness exploration. Finally, we conclude the paper in Section 6.

\section{Related Work}
\label{sec2}
\subsection{Adversarial Attacks on GNNs}
With the increasing popularity of GNNs, adversarial attacks on them have attracted significant attention in recent years. Gradient-based adversarial attack methods leverage back-propagation to compute the gradient of the loss function with respect to the input data. This allows the identification and perturbation of the links or node features that have the greatest impact on the performance of GNNs. For example, Chen et al. \cite{chen2018fast} proposed the Fast Gradient Attack (FGA), which generates adversarial examples using the gradient information of GCN. Z{\"u}gner et al. \cite{zugner2019adversarial} proposed the poisoning attack method Mettack, which treats the graph structure matrix as a hyperparameter and computes the gradient of the attacker's loss with respect to it. Wu et al. \cite{wu2019adversarial} introduced IG-JSMA, a gradient-guided adversarial attack method that computes the gradients of the prediction score concerning the adjacency and feature matrices, which are used to perturb links or features. Fang et al. \cite{lin2023exploratory} proposed EpoAtk, an approach that enhances gradient-based perturbations on graphs through an exploratory strategy.

Many adversarial attack methods based on reinforcement learning have also been proposed. Dai et al. \cite{dai2018adversarial} introduced RL-S2V, a reinforcement learning-based attack method that learns the graph attack policy using only the prediction labels of the target classifier. Sun et al. \cite{sun2020adversarial} proposed a deep hierarchical reinforcement learning-based method, NIPA, that models the key steps of node injection attacks through a Markov decision process. Ju et al. \cite{ju2023let} presented the $G^2$A2C attack method, which leverages reinforcement learning to inject highly malicious nodes while operating under extremely limited attack budgets. Chen et al. \cite{chen2024single} proposed $G^2$-SNIA, a reinforcement learning framework that employs proximal policy optimization for single-node injection attacks. 

In addition to the aforementioned approaches, several optimization-based adversarial attack methods have been proposed. Wang et al. \cite{wang2019attacking} introduced a threat model for characterizing the attack surface of collective classification methods by manipulating the graph structure through optimization modeling. Geisler et al. \cite{geisler2021robustness} proposed two sparsity-aware first-order optimization attacks. Zou et al. \cite{zou2021tdgia} proposed the Topological Defective Graph Injection Attack (TDGIA) method, which designs a smooth feature-optimization objective to generate the features of the injected nodes. Sharma et al. \cite{sharma2023node} presented NICKI, an optimization-based node injection method that jointly optimizes the features and edges of the injected nodes. 

\subsection{Adversarial Robustness of GNNs}
Extensive efforts have been made to improve the adversarial robustness of GNNs, which can generally be categorized into three classes: preprocessing, adversarial training, and robust model design methods. Specifically, preprocessing-based methods aim to remove or weaken the adversarial perturbations before model training so that the resulting graphs resemble the original ones. Wu et al. \cite{wu2019adversarial} utilized Jaccard similarity to assess the likelihood of links and improved the adversarial robustness by removing links connecting significantly different nodes. Entezari et al. \cite{entezari2020all} found that only the high-rank singular values of graphs are affected by Nettack and introduced a low-rank approximation method to eliminate adversarial perturbations. Zhu et al. \cite{zhu2023focusedcleaner} presented a poisoned graph sanitizer to effectively identify poisoned samples injected by attackers through bi-level structure learning and victim node detection.

Adversarial training-based methods aim to defend against adversarial attacks by augmenting the training data with carefully crafted adversarial examples. Feng et al. \cite{feng2019graph} introduced GraphAT, an adversarial training method with dynamic regularization, to enhance the model's adversarial robustness and generalization. Deng et al. \cite{deng2023batch} proposed BVAT, a batch virtual adversarial training method based on GCNs, which aims to smooth the classifier's output distribution. By solving the min-max problem, Xu et al. \cite{xu2019topology} proposed an optimization-based adversarial training method that is robust to both optimization-based and greedy search-based topological attacks. Sun et al. \cite{sun2019virtual} applied virtual adversarial training to the supervised losses of GCNs to enhance their generalization performance. 

Robust model design-based methods leverage the characteristics of adversarial perturbations to eliminate their negative effects. Feng et al. \cite{feng2020graph} proposed a semi-supervised learning framework called GRAND, which includes graph data expansion and consistent regularization. Jin et al. \cite{jin2020graph} proposed a general framework, ProGNN, which jointly learns a structural graph and robust model from the perturbed graph. Ioannidis et al. \cite{ioannidis2020tensor} introduced TGCN, a tensor-based semi-supervised GCN. Liu et al. \cite{liu2021elastic} developed Elastic GNNs, a series of architectures that integrate message-passing schemes into DNNs to improve adversarial robustness. Geisler et al. \cite{geisler2021robustness} proposed a robust and differentiable aggregation function, Soft Median, which significantly reduces memory overhead and enables effective defense on large-scale graphs. Wang et al. \cite{wang2024trustguard} proposed TrustGuard, a robust GNN-based model for accurate trust evaluation that supports trust dynamics. Wang et al. \cite{wang2023fl} proposed FL-GNNs, a general framework to improve robustness by jointly learning feature representations and denoised node features. This approach suppresses feature noise during training and can be integrated into various GNN architectures. 

\subsection{Explainability Methods for GNNs}
The explainability of GNNs is critical for understanding their underlying mechanisms and improving their adversarial robustness. Recent studies have increasingly focused on enhancing the interpretability of GNNs. For example, Ying et al. \cite{ying2019gnnexplainer} introduced GNNExplainer, a generic method for explaining the predictions of any GNN-based model. This approach identifies a compact subgraph and a minimal subset of node features that are crucial for a GNN's prediction. Yuan et al. \cite{yuan2020xgnn} proposed the XGNN approach to explain GNNs by training a graph generator to produce graph patterns that maximize the model's prediction confidence. Yuan et al. \cite{yuan2021on} proposed SubgraphX, a method for interpreting GNNs by identifying significant subgraphs using Monte Carlo tree search. Huang et al. \cite{huang2022graphlime} proposed the generic GNN explanation framework, GraphLIME, which locally learns a nonlinear and interpretable model. Zhang et al. \cite{zhang2024expressive} presented a unified theoretical framework for analyzing the expressive power of GNNs from the perspectives of feature embedding and topology representation, providing valuable insights for understanding and enhancing adversarial robustness. 

To explore the factors affecting the adversarial vulnerability of GNNs, Chen et al. \cite{chen2021understanding} attributed the success of adversarial attacks on GCNs to a non-robust aggregation scheme (i.e., the weighted mean). Zhu et al. \cite{zhu2022how} found that adversarial attacks on GNNs are primarily due to increased graph heterogeneity. By analyzing popular defense methods, Mujkanovic et al. \cite{mujkanovic2022defenses} found that most defense methods demonstrate minimal to no improvement compared to the undefended baseline. However, there are few studies on the underlying mechanisms of GNNs' vulnerability, and existing studies remain fragmented. In addition, Dai et al. \cite{dai2024comprehensive} presented a comprehensive survey on trustworthy GNNs from the perspectives of privacy, robustness, fairness, and explainability, emphasizing that explainability not only aids interpretation but also reveals biased decisions and adversarial vulnerabilities. Their findings highlight the importance of aligning interpretability with robustness to develop more trustworthy GNNs.

In contrast to the aforementioned methods, our goal is to comprehensively understand the underlying mechanisms of adversarial attacks and to discover the key factors that influence model vulnerability, thereby supporting the design of robust GNNs.

\section{Problem Definition and Preliminaries}
\label{sec3}
\subsection{Notations}
Let ${\mathcal G} = \left( {{\mathcal{V}},{\mathcal E}} \right)$ represent a graph, where ${\mathcal{V}} = \left\{ {{v_1},\ldots,{v_N}} \right\}$ is a set of $N$ nodes, ${\mathcal E} = \left\{ {{e_1},\ldots,{e_K}} \right\}$ is a set of $K$ edges, and ${e_{i,j}} = \left( {{v_i},{v_j}} \right)$ represents an edge between nodes $v_i$ and $v_j$. Formally, we denote the adjacency matrix of ${\mathcal G}$ as $\mathbf{A} \in {\mathbb{R}^{N \times N}}$, where ${{\mathbf{A}}_{ij}} = 1$ if ${v_i}$ and ${v_j}$ are connected in ${\mathcal G}$; otherwise, ${{\mathbf{A}}_{ij}} = 0$. We denote the feature matrix as $\mathbf{X} \in {\mathbb{R}^{N \times D}}$, where $D$ is the dimension of the feature vector. Thus, the graph can be represented as ${\mathcal G} = \left( {{\mathbf{A}},{\mathbf{X}}} \right)$.

\subsection{Problem Definition}
\noindent \textbf{{\emph{Definition 1 (Adversarial Attack on Graphs). }}} For an original graph ${\mathcal G}$, the attacker manipulates the graph structure or node features to generate an adversarial graph $\hat {\mathcal G} =\left( {\hat {\mathbf{A}},\hat {\mathbf{X}}} \right)$ by adding imperceptible adversarial perturbations $ \delta $. This will lead the model $f\left( \cdot  \right)$ to make incorrect predictions, while minimizing the perturbation between $\mathcal G$ and $\hat {\mathcal G}$.
\begin{equation}
	\arg \min_\delta \|\delta\| \; \text{subject to} \; f(\hat{\mathcal G}) \ne f(\mathcal G)
\end{equation}

In this paper, we focus only on the adversarial attacks that perturb the graph structure, i.e., $\hat {\mathcal G} = \left( {{\mathcal{V}},\hat {\mathcal E}} \right),\;\hat {\mathcal E} = {\mathcal E} + \delta$.

\noindent \textbf{{\emph{Definition 2 (Adversarial Defense). }}} For the perturbed graph $\hat {\mathcal G}$, the goal of adversarial defense is to minimize the loss function $\mathcal{L}_{atk}$ of the attacked model so that it is as consistent as possible with the loss of the clean model. A model $f(\cdot)$ is considered robust if it can maintain its performance under adversarial attacks.
\begin{equation}
	f(\hat{\mathcal G}) \to y \; \text{and} \; f(\mathcal G) \to y
\end{equation}

\noindent \textbf{{\emph{Definition 3 (Robust Explanation of GNNs). }}} Let $f\left(  \cdot  \right)$ denote the model that needs to be explained. Given a perturbed graph $\hat {\mathcal G}$ generated by adversarial attack methods, the goal of robust explanation of GNNs is to identify the key factors ${\mathcal{P}^*}$ that influence the performance of GNNs from the set of all possible options $\mathcal{P}$.
\begin{equation}
	{\mathcal{P}}^* = \left\{ {p\;|\;f{(\hat {\mathcal G}|p)}},\;p \in {\mathcal{P}}\right\}
\end{equation}
where $p$ denotes an element of the option set $\mathcal{P}$. By evaluating $f(\hat{\mathcal{G}} | p)$ for each $p$, we identify those that significantly affect adversarial robustness and include them in the set $\mathcal{P}^*$. 

\subsection{Adversarial Attacks on GNNs}
According to the goal of the attackers, adversarial attacks on GNNs can be categorized as untargeted, targeted, and random attacks, in which the representative methods are Mettack \cite{zugner2019adversarial}, Nettack \cite{zugner2018adversarial}, and Random Attack \cite{malik2017robustness}, respectively.

\subsubsection{Mettack}
To reduce the overall classification accuracy, Mettack treats the graph structure as a hyperparameter and computes meta-gradients based on a bilevel optimization framework. The method is formulated as:
\begin{equation}
	\nabla _\mathcal{G}^{meta}: = {\nabla _\mathcal{G}}{\mathcal{L}_{atk}}\left( {{f_{\theta ^*}}\left( \mathcal{G} \right)} \right),\theta ^* = {{opt}_\theta }\left( {{\mathcal{L}_{train}}\left( {{f_\theta }\left( \mathcal{G} \right)} \right)} \right)
\end{equation}
where ${\mathcal{L}_{atk}}$ is the loss function for optimization, $\theta$ is the parameter of the surrogate model, ${\mathcal{L}_{train}}$ is the training loss function, and $opt\left(  \cdot  \right)$ is a differentiable optimization process. Using the obtained meta-gradient, the attacker updates the graph and obtains the final poisoning graph.

\subsubsection{Nettack}
Nettack performs attacks based on the loss of GNNs by adopting a two-layer GCN as a surrogate model. The activation function $\sigma$ is replaced by the identity function, such that $\mathbf{W} = \mathbf{W}^{(1)} \cdot \mathbf{W}^{(2)}$, leading to:
\begin{equation}
	\mathbf{Z} = {\mathrm{softmax}}\max \left( {{{\mathbf{A}}^2}{\mathbf{X}}{\mathbf{W}}} \right)
\end{equation}

Since the normalization factor in the softmax function does not influence the prediction category, the surrogate model can be simplified as ${{{\mathbf{A}}^2}{\mathbf{X}}{\mathbf{W}}}$. Accordingly, the surrogate loss can be defined as follows:
\begin{equation}
	{\mathcal{L}_\mathcal{S}}(\mathbf{A},\mathbf{X};\mathbf{W},{v_0}) = \mathop {\max }\limits_{c \ne {c_{old}}} {\left[ {\hat {\mathbf{A}}\mathbf{X}\mathbf{W}} \right]_{{v_0}c}} - {\left[ {\hat {\mathbf{A}}\mathbf{X}\mathbf{W}} \right]_{{v_0}c}}_{old}
\end{equation}
where $\mathcal{S}$ denotes the surrogate model. Then, scoring functions $\mathcal{S}_{struct}$ and $\mathcal{S}_{feat}$ are introduced to evaluate the loss of the perturbation model.
\begin{align}
	{\mathcal{S}_{struct}}\left( {e;\mathcal{G},{v_0}} \right): = {\mathcal{L}_s}\left( {\mathbf{A}',\mathbf{X};\mathbf{W},{v_0}} \right)\\
	{\mathcal{S}_{feat}}\left( {f;\mathcal{G},{v_0}} \right): = {\mathcal{L}_s}\left( {\mathbf{A},\mathbf{X}';\mathbf{W},{v_0}} \right)
\end{align}	

Subsequently, the perturbation that maximizes the scoring functions is selected and applied to the graph until a predefined perturbation budget is reached. 

\subsubsection{Random Attack}
Random attack involves randomly deleting or inserting a small number of nodes or edges from the clean graph under different perturbation rates. Compared to other attacks, it does not require any prior knowledge about the target model and training data, and has the lowest computational cost.

\section{Robustness Exploration Framework}
\label{sec4}
\subsection{Framework}
\begin{figure*}[htbp]
	\centering
	\includegraphics[width=0.95\textwidth]{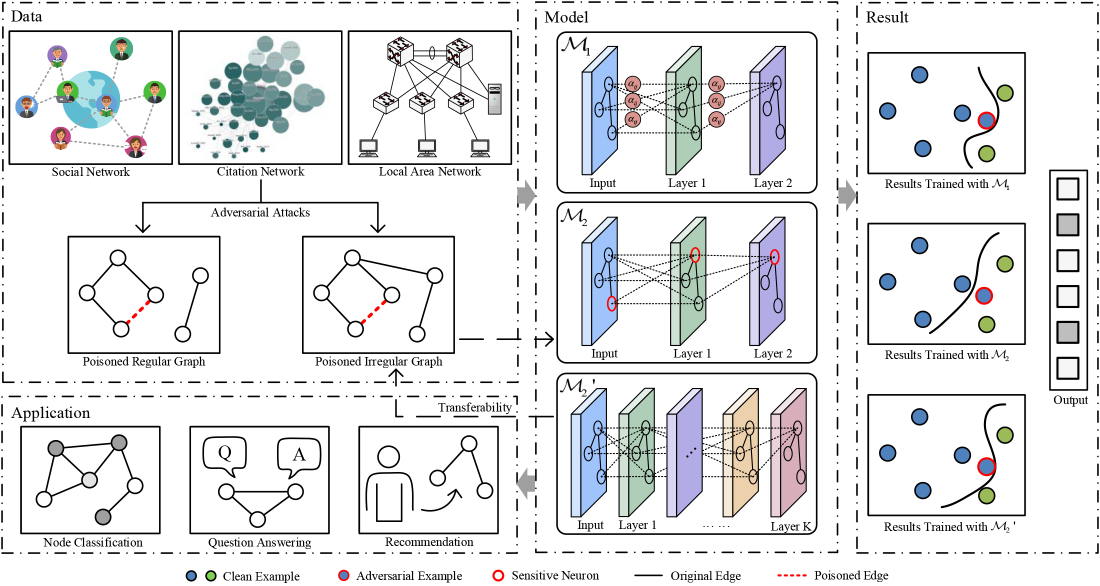}
	\caption{Adversarial robustness exploration framework of GNNs. (1) The adversarial robustness analysis of models trained on regular and irregular graph patterns. (2) The adversarial robustness analysis of models with different model architectures (${{\mathcal M}_1}$ vs. ${{\mathcal M}_2}$), as well as the impact of adversarial attacks on sensitive neurons. (3) The adversarial robustness analysis of models with different model capacities (${{\mathcal M}_2}$ vs. ${{\mathcal M}_2}'$). }
	\label{fig1}
\end{figure*}
This section presents the general framework for investigating the adversarial robustness of GNNs, as shown in Fig.~\ref{fig1}. The framework is designed to reveal how various factors---including graph patterns, model architecture, and model capacity---influence the robustness of GNNs against adversarial perturbations. In the left section of Fig. Fig.~\ref{fig1}, we focus on how structural patterns of input graphs affect the model's adversarial robustness. Specifically, GNNs are trained on both regular and irregular graphs derived from real-world domains such as social networks, citation networks, and local area networks. These graphs are perturbed with adversarial edges (highlighted in red), and the resulting structural differences directly influence the learned decision boundaries and their robustness against adversarial perturbations. The middle section compares two model architectures (${{\mathcal M}_1}$ vs. ${{\mathcal M}_2}$) to examine how architectural design choices affect adversarial robustness. This comparison goes beyond overall performance and focuses on the stability of internal representations under adversarial perturbations. We also conduct a neuron-level analysis to identify sensitive components that contribute to GNNs' vulnerability. Furthermore, by comparing ${{\mathcal M}_2}$ with its deeper variant ${{\mathcal M}_2}'$, we explore the impact of model capacity by analyzing how increased depth affects adversarial robustness. The right section presents classification results under these settings. It shows that adversarial examples (circled in red) interact differently with the decision boundaries of each model, indicating that adversarial robustness is closely related to how well the model separates classes under perturbations. Additionally, the framework considers the transferability of adversarial examples across models, emphasizing that adversarial examples generated from one model can often fool other models with different architectures or capacities.

\subsection{Graph Patterns}
The organization of real-world graphs typically consists of both regular and irregular components, and only the former can be explained and modeled. Hence, the significance of graph patterns directly influences the training of target models. In practice, the structural regularity of a graph can be assessed by the consistency of its structural features before and after randomly removing a small subset of links. Consequently, the structural regularity of graphs coincides with our ability to predict missing links in them \cite{lu2015toward}: 
\begin{equation}
	\sigma_c = \left| {\mathcal E}^L \cap \Delta {\mathcal E} \right| /  \Delta {\mathcal E}
\end{equation}
where ${\mathcal E}^L$ is the set of top-L predicted links and $\Delta {\mathcal E}$ is the set of randomly removed links. Thus, the graph pattern is a crucial factor worth investigating in our analysis of the adversarial robustness of GNNs.

\textbf{Intuition.} In adversarial attack scenarios, we assume that adversarial attacks disrupt the structural patterns of the input graphs, thus increasing their irregularity. Hence, adversarial perturbations can be viewed as part of the irregular components within graphs. As the models trained exclusively on regular graphs that lack any adversarial examples have difficulty recognizing adversarial perturbations, the intrinsic complexity of GNNs raises a pivotal question: how does the degree of structural regularity in a graph influence the robustness of models against adversarial attacks? We assume that GNNs trained with either regular or irregular graphs will demonstrate varying degrees of robustness. Specifically, GNNs trained solely on highly regular graphs (as depicted in the ``Poisoned Regular Graph'' in Fig. \ref{fig1}) are more susceptible and tend to misclassify the target node when subjected to adversarial attacks. Conversely, GNNs trained on irregular graphs (as illustrated in the ``Poisoned Irregular Graph'' in Fig. \ref{fig1}) tend to exhibit greater adversarial robustness and are thus more likely to make accurate predictions when faced with adversarial attacks.

\subsection{Model Architecture}
Since the seminal GCN \cite{kipf2016semi} model was introduced for semi-supervised node classification, a multitude of other notable models, such as GAT \cite{velickovic2017graph} and GraphSAGE \cite{hamilton2017inductive}, have subsequently been developed. Building on this foundation, the GNNs literature has witnessed a surge in the development of numerous innovative models, each designed to address a broad spectrum of task-specific scenarios, learning paradigms, and complex data characteristics. Generally, GNNs are frequently designed through a trial and error approach, and there exists a high degree of independence among the internal mechanisms of different GNNs, accompanied by a lack of evolutionary development relationships between various model architectures. Meanwhile, the research community lacks systematic exploration and analysis on the adversarial robustness of typical GNNs, and the architectural design of GNNs lacks a clear research direction. It is of great significance to provide principled guidance for the architectural design of GNNs based on the robustness of individual models when facing adversarial attacks. Thus, exploring the performance of representative GNNs under adversarial attacks is meaningful for understanding the adversarial robustness properties of model architecture and guiding the design of robust models. 

In GNNs with various architectures, neurons are fundamental units that perform nonlinear transformations through activation functions to produce the final output \cite{dou2023understanding}. Since the model output arises from the joint action of neurons across hidden layers, adversarial attacks can be traced to perturbations that significantly alter the activation values of individual neurons, i.e., through imperceptible changes that disrupt the neuron responses. In other words, adversarial attacks occur when neurons are inappropriately activated, resulting in erroneous hidden representations and ultimately inaccurate outputs. Therefore, exploring the impact of adversarial attacks on
neurons provides clearer insight into the adversarial robustness of deep
learning models. 

\textbf{Intuition.} Different model architectures of GNNs have distinct internal mechanisms, and an appropriate model architecture can generate robust representations, thereby improving model performance. As illustrated in the middle section of Fig. \ref{fig1}, models with different architectures (${{\mathcal M}_1}$ vs. ${{\mathcal M}_2}$) exhibit substantially different decision boundaries in the embedding space, resulting in varying degrees of robustness against adversarial perturbations. Inspired by the design space for GNNs \cite{you2020design}, existing works have primarily focused on specific designs of GNNs rather than the model design space, limiting the discovery of robust GNNs. Therefore, it is necessary to systematically study the architecture of GNNs against adversarial attacks. In addition, from the perspective of model architecture, it is still unclear how adversarial attacks affect GNNs, for example, whether they slightly affect a large number of neurons or significantly impact a small number of neurons. Overall, we expect this investigation to provide a principled approach and foundational model architecture for robust model design. This issue is particularly critical considering the large model space of GNNs and the diversity of adversarial attack methods, as re-exploring all possible combinations of architecture components is prohibitively expensive.

\subsection{Model Capacity}
Model capacity refers to a model's ability to fit various functions and effectively map inputs to outputs. In machine learning, a model with insufficient capacity may struggle to adequately characterize the training data, while a model with excessive capacity may simply memorize the training data. In other words, models with different capacities will produce different prediction results. Taking GCN as an example, increasing the model's capacity by adding more convolutional layers allows it to capture a broader range of graph information, thereby enhancing its representational power and predictive performance. Consequently, when dealing with input graphs that contain adversarial examples, GNNs with varying capacities will exhibit varying levels of performance. Therefore, we investigate the adversarial robustness of GNNs from the perspective of model capacity.

\textbf{Intuition.} In the context of adversarial attacks, model capacity plays a crucial role in determining a model's robustness. To accurately characterize and effectively resist adversarial perturbations, models often require greater capacity than that needed to correctly classify benign examples. In this study, we define the model capacity of GNNs by the number of hidden layers, with deeper layers corresponding to higher capacity. As shown in the middle and right sections of Fig. \ref{fig1}, lower-capacity models (${{\mathcal M}_2}$) tend to form simpler decision boundaries, making samples near the margin (i.e., close to the decision boundary) more vulnerable to adversarial perturbations that push them across class boundaries. In contrast, higher-capacity models (${{\mathcal M}_2}'$) learn more complex and expressive boundaries that are more effective at mitigating such perturbations, resulting in improved adversarial robustness. Moreover, according to \cite{tian2021detecting}, the adversarial robustness of deep learning models is closely related to the complexity of their decision boundaries, with adversarial perturbations typically directed towards high-curvature regions of these boundaries. As different model capacities lead to varying decision boundary complexities, studying GNNs' adversarial robustness from the perspective of model capacity offers a promising direction for better understanding and improving adversarial robustness. 

\subsection{Evaluation Metrics}
\label{sec_Evaluation Metrics}
To evaluate the adversarial robustness of GNNs, we adopt classification accuracy under adversarial attacks as the main metric, with higher classification accuracy indicating greater adversarial robustness. In addition, we introduce the decision surface and the adversarial transferability rate (ATR) to further assess the robustness and transferability of the adversarial examples.

\subsubsection{Decision Surface} 
\begin{figure}[h]
	\centering
	\begin{subfigure}[b]{0.48\columnwidth} 
		\centering
		\includegraphics[width=\columnwidth]{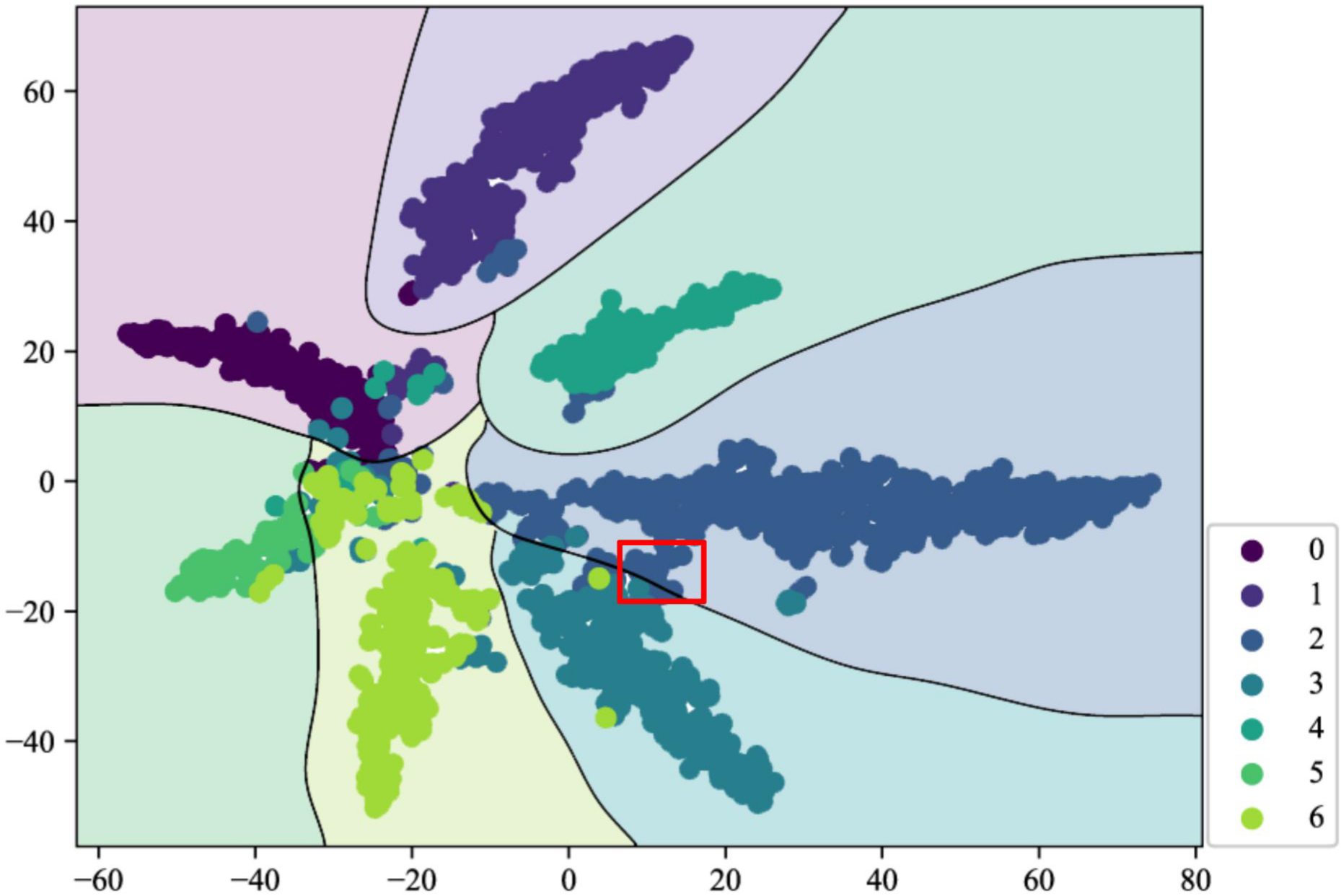}
		\caption{}
	\end{subfigure}
	\hfill
	\begin{subfigure}[b]{0.48\columnwidth} 
		\centering
		\includegraphics[width=\columnwidth]{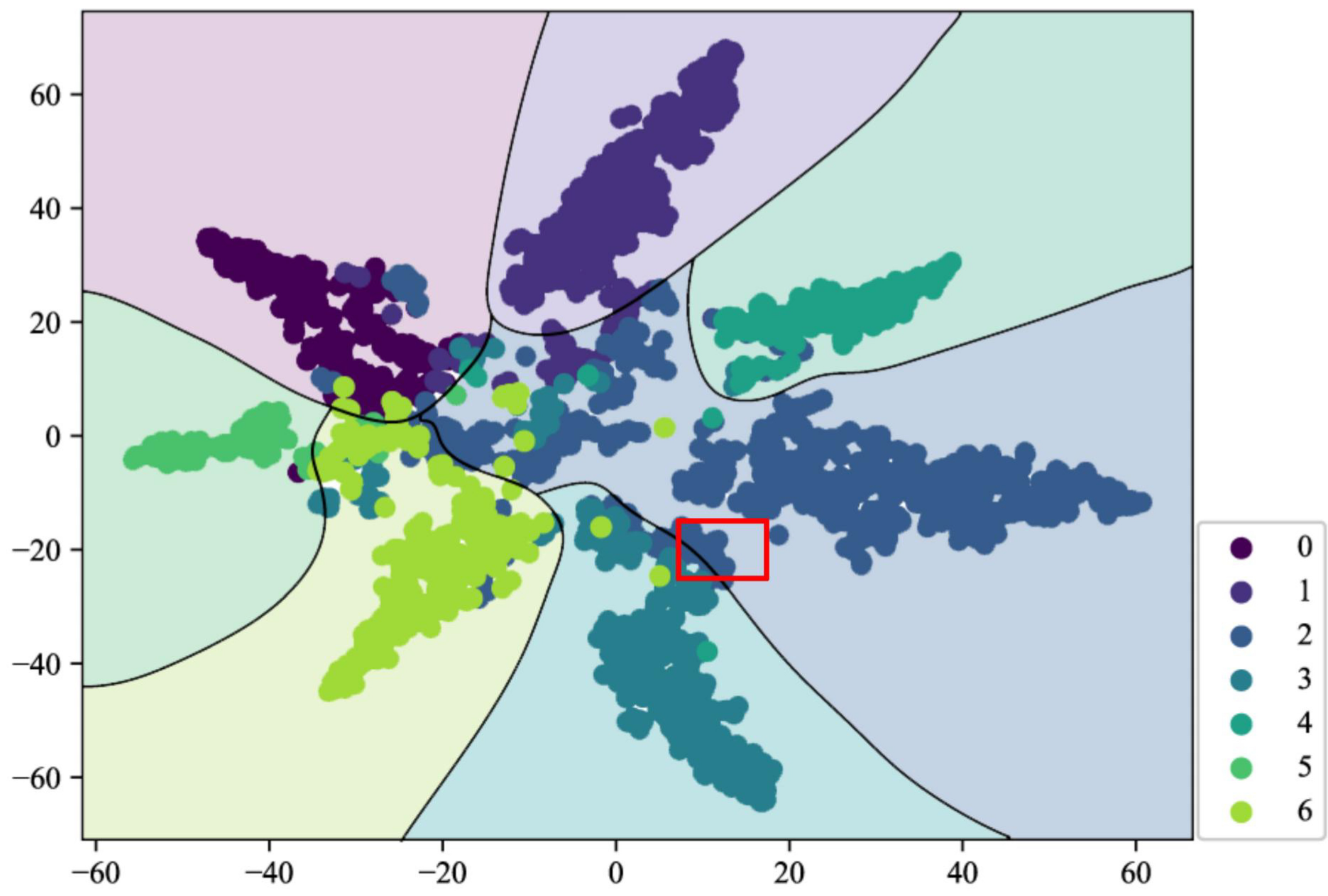}
		\caption{}
		\label{fig2_boundary_b}
	\end{subfigure}
	\begin{subfigure}[b]{\columnwidth} 
		\centering
		\includegraphics[width=0.93\columnwidth, height=3.1cm]{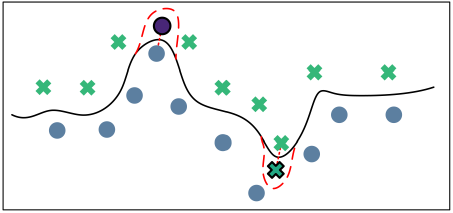}
		\caption{}
		\label{fig2_boundary_c}
	\end{subfigure}
	\caption{Illustration of GCN's decision boundaries on Cora. (a) Decision boundaries of GCN. (b) Decision boundaries of GCN under Mettack with 5\% perturbation. (c) Decision boundaries before and after adversarial attack.}
	\label{fig2_boundary}
\end{figure}

The decision boundary is a surface that separates the data points from different classes. For a classifier ${\mathcal F}$, the decision boundary between adjacent classes $i$ and $j$ is defined as:
\begin{equation}
	\mathbb{B} = \left\{ z \in \mathbb{S}: {\mathcal F}_i(z) - {\mathcal F}_j(z) = 0 \right\}
\end{equation}
where ${{\mathcal F}_i}\left( z \right)$ and ${{\mathcal F}_j}\left( z \right)$ are the ${i^{th}}$ and ${j^{th}}$ components of ${\mathcal F}\left( z \right)$, corresponding to classes ${i}$ and ${j}$, respectively. Accordingly, for a graph ${\mathcal G}$ with decision space ${\mathbb{S}}$, the GNN model ${\mathcal F}$ partitions ${\mathbb{S}}$ into $r$ decision regions, denoted as ${s_1},{s_2},\ldots,{s_r}$. The decision boundaries between adjacent classes in ${\mathcal G}$ can be expressed as: 
\begin{equation}
	\mathbb{DB}_{i,j} = \left\{ \mathbf{h}_v \in \mathbb{S}: {\mathcal F}_i(\mathbf{h}_v) = {\mathcal F}_j(\mathbf{h}_v),\ i \ne j \right\}
	\label{eq_dbij}
\end{equation}
where $\mathbf{h}_v$ denotes the embedding of node $v$ in space $\mathbb{S}$. Fig. \ref{fig2_boundary} presents a representative example of adversarial robustness analysis based on decision boundaries, where the regions corresponding to each category represent decision regions and the black lines separating them indicate the decision boundaries. After applying slight and imperceptible adversarial perturbations (Fig. \ref{fig2_boundary_b}), the decision boundaries exhibit noticeable deviations, resulting in misclassifications. To better understand these deviations, Fig. \ref{fig2_boundary_c} shows a detailed illustration of the highlighted region in Fig. \ref{fig2_boundary_b}. A slight adversarial perturbation causes the decision boundary to shift from its original state (black solid line) to a perturbed state (red dashed line), leading node representations near the boundary (light blue circle and light green cross) to cross it and deviate from their original class regions. In contrast, if the decision boundary remains largely unchanged under adversarial attacks, the model is considered robust. 

Intuitively, the difference between the predictions for different classes of samples should be as large as possible, and a robust model should exhibit high prediction confidence. However, the model ${\mathcal F}$ with the decision boundaries defined in Equation \eqref{eq_dbij} cannot effectively form an explicit decision boundary. Therefore, according to \cite{fawzi2018empirical}, for the node representation ${\mathcal Z}\left( {\mathcal G} \right)$ obtained before the softmax layer of model ${\mathcal F}$, the decision surface can be defined as:
\begin{equation}
	{\mathcal S}\left( {\mathcal G} \right) = {\mathcal Z}{\left( {\mathcal G} \right)_t} - \max \left\{ {{\mathcal Z}{{\left( g \right)}_i},i \ne t} \right\}
\end{equation}
where $ t $ denotes the ground-truth class label of $ \mathcal{G} $.

In neural networks, the loss function quantifies the variation in model output to input perturbations and is used to evaluate the target model. Accordingly, by treating ${\mathcal S}\left( {\mathcal G} \right)$ as the loss function, given an input graph ${\mathcal G}$ for GNNs, the perturbed graphs around it can be constructed and evaluated to obtain the corresponding loss values: 
\begin{equation}
	{\mathcal V}\left( {{\mathcal G},\alpha ,\beta } \right) = {\mathcal S}\left( {{\mathcal G} + (\alpha \cdot P\left( {\mathcal V} \right),\beta \cdot P\left( {\mathcal E} \right))} \right)
\end{equation}
where $\alpha $ and $\beta $ represent the degree of node perturbation ${P\left( {\mathcal V} \right)}$ and edge perturbation ${P\left( {\mathcal E} \right)}$, respectively. A larger value of ${\mathcal S}\left( {\mathcal G} \right)$ indicates greater adversarial robustness.

\subsubsection{Adversarial Transferability Rate}
Adversarial examples generated from a source model can often fool a target model due to their transferability. To assess the security risks of adversarial transferability, we define the ATR, using adversarial examples generated specifically for the target model as a baseline:
\begin{equation}
	ATR = (Acc_{\text{transfer}} - Acc_{\text{specific}})/{Acc_{\text{specific}}}
	\label{eq_atr}
\end{equation}
where $Acc_{\text{transfer}}$ denotes the accuracy of the target model attacked by the adversarial examples transferred from the source model, and $Acc_{\text{specific}}$ denotes the accuracy of the target model attacked by the adversarial examples generated specifically for the target model itself. Specifically, $ATR = 0$ indicates that transfer-based attacks have the same effect as local-based attacks, whereas $ATR < 0$ denotes that transfer-based attacks have a more significant effect on the accuracy of the target model than local-based attacks.

\section{Experiments: Robustness Exploration}
\label{sec5}
In this section, we investigate the effects of graph patterns, model architecture, and model capacity on the adversarial robustness of GNNs, and summarize key findings along with practical guidelines for enhancing robustness.

\subsection{Experimental Settings}
\subsubsection{Datasets}
We conducted experiments on both synthetic and real-world datasets. For synthetic datasets, we adopted the Lancichinetti Fortunato Radicchi (LFR) benchmark network generation algorithm \cite{lancichinetti2008benchmark} to produce graphs with varying structural regularities. 
For real-world datasets, we utilize four widely used benchmark datasets: Cora \cite{mccallum2000automating}, Citeseer \cite{giles1998citeseer}, PubMed \cite{sen2008collective}, and Amazon Photo \cite{shchur2018pitfalls}. Specifically, (1) Cora is a citation network with 2,485 papers in 7 classes, represented by 1,433-dimensional binary word vectors; (2) Citeseer is a citation network with 2,110 papers in 6 classes and 3,668 citation links, represented by binary word vector features; (3) PubMed is a citation network of 19,717 diabetes-related papers in 3 classes, connected by 44,338 links and represented by 500-dimensional TF-IDF vectors; (4) Amazon Photo is a co-purchase network of 7,650 products in 8 categories, represented by 745-dimensional bag-of-words features and connected through 238,162 co-purchase edges.

\subsubsection{Models and Attacks}
We evaluated the adversarial robustness of representative classical GNNs across five architectural categories. For Convolutional-GNNs, GCN \cite{kipf2016semi} uses a first-order spectral approximation for efficient aggregation; ChebNet \cite{defferrard2016convolutional} extends this with Chebyshev polynomials for multi-hop propagation; and SGC \cite{wu2019simplifying} simplifies GCN by collapsing weights into a linear scheme. For Attention-GNNs, GAT \cite{velickovic2017graph} applies self-attention to assign adaptive weights to neighbors. In Diffusion-GNNs, APPNP \cite{gasteiger2018predict} decouples transformation and propagation via personalized PageRank, while AirGNN \cite{liu2021graph} enhances this with adaptive residual propagation for better feature preservation. For Sampling-GNNs, GraphSAGE \cite{hamilton2017inductive} uses sampling-based aggregation with learnable functions. Finally, for Isomorphism-GNNs, GIN \cite{xu2018powerful} employs injective aggregators and MLPs to improve representation power, especially for distinguishing non-isomorphic graphs.

Regarding adversarial attacks, we adopt three widely used methods: Mettack, Nettack, and Random Attack. For Mettack and Random Attack, we apply structural perturbations with perturbation rates ranging from 2\% to 10\% (i.e., 2\%, 4\%, 5\%, 6\%, 8\%, and 10\%). For Nettack, we perturb each target node with either 2.0 or 4.0 structural modifications. In the experiments, we focus on poisoning attacks, where the target models are retrained on the perturbed graphs. This setup presents greater challenges for defenders and better simulates real-world deployment scenarios. 

\subsubsection{Parameter Settings}
All experiments are conducted over ten independent runs, and the reported results are averaged accordingly to ensure statistical reliability. The models are implemented in PyTorch version 1.8.2 and executed on an NVIDIA RTX 3080Ti GPU equipped with CUDA 11.1. The ReLU activation function is used throughout, the dropout rate is set to 0.5 to prevent overfitting, and all models are consistently trained for a total of 200 epochs.

\subsection{Effect of Graph Patterns}
\subsubsection{Structural Regularity of Graph}
\begin{figure}[htbp]
	\centering
	\includegraphics[width=\columnwidth]{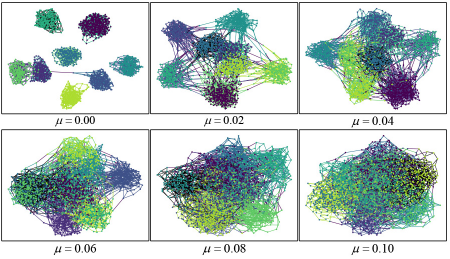}
	\caption{Artificial graphs with different structural regularities.} 
	\label{fig3_lfr}
\end{figure}

To illustrate the impact of different structural patterns in graph data on the adversarial robustness of GNNs, we generate synthetic graphs with various structural regularities based on the community structures observed in real systems. Specifically, we hypothesize that graphs with a strong community structure exhibit high structural regularity. The LFR network generation algorithm is adopted to create synthetic graphs with different structural regularities, as shown in Fig. \ref{fig3_lfr}. The LFR benchmark provides a flexible set of parameters, enabling the generation of diverse yet realistic graphs, where the mixed parameter $\mu$ quantifies the regularity of the graphs, and is defined as:
\begin{equation}
	\mu = {K_c} / {N_e}
\end{equation}
where ${N_e}$ represents the total number of links for the current node and ${K_c}$ is the number of links to nodes in other communities. In other words, each node shares $1-\mu$ of its links with the other nodes in its community and $\mu$ with the other nodes in the network. A larger $\mu$ indicates more inter-community connections and lower structural regularity. When $\mu$ is zero, there are no edges between communities, resulting in a perfectly pure and regular graph. In Fig. \ref{fig3_lfr}, we set $\mu$ as 0.00, 0.02, 0.04, 0.06, 0.08 and 0.10, indicating a transition from strong to weak structural regularity.

\begin{table*}[h]
	\centering
	\caption{Accuracy of different regularity graphs for GCN under three typical adversarial attacks. \textbf{Bold} numbers indicate the best results of different regularities.}
	\renewcommand{\arraystretch}{1.2} 
	\setlength{\tabcolsep}{10pt} 
	\begin{tabular*}{0.9\textwidth}{@{\extracolsep{\fill}} l *{6}{p{1.1cm}}}
		\toprule
		\multirow{2}{*}{Method (Ptb Rate/Num)} & \multicolumn{6}{c}{$\mu$} \\ \cline{2-7} 
		& 0.00 & 0.02 & 0.04 & 0.06 & 0.08 & 0.10 \\ 
		\midrule
		Mettack(0.00) & 0.7186 & 0.7550 & \textbf{0.7693} & 0.7639 & 0.7504 & 0.7492 \\
		Mettack(0.05) & 0.7184 & 0.7427 & 0.7550 & \textbf{0.7748} & 0.7498 & 0.7489 \\
		Mettack(0.10) & 0.7049 & 0.7064 & 0.6411 & \textbf{0.7712} & 0.6318 & 0.6296 \\
		\hline
		Nettack(0.0) & 0.7108 & \textbf{1.0000} & 1.0000 & 1.0000 & 1.0000 & 1.0000 \\
		Nettack(2.0) & 0.5012 & \textbf{0.6398} & 0.6169 & 0.6120 & 0.5699 & 0.5675 \\
		Nettack(4.0) & 0.1470 & 0.1211 & 0.1048 & \textbf{0.9319} & 0.7982 & 0.7108 \\
		\hline
		Random Attack(0.00) & 0.7181 & 0.7541 & 0.7582 & \textbf{0.7635} & 0.7509 & 0.7507 \\
		Random Attack(0.05) & 0.7160 & 0.7538 & 0.7573 & \textbf{0.7627} & 0.7508 & 0.7505 \\
		Random Attack(0.10) & 0.7153 & 0.7509 & 0.7564 & \textbf{0.7618} & 0.7498 & 0.7481 \\ 
		\bottomrule
	\end{tabular*}
	\label{table1_datapattern}
\end{table*}

The classification accuracy of GCN across different graph patterns under three representative adversarial attack methods is summarized in Table \ref{table1_datapattern}. The header row indicates the mixing parameter $\mu$, with values increasing from left to right, reflecting a gradual decrease in the structural regularity of the graphs. The header column lists the adversarial attack methods along with their corresponding perturbation rates or numbers. As shown in Table \ref{table1_datapattern}, the classification accuracy is lowest when $\mu = 0.00$, peaks at $\mu = 0.06$ (except for a local maximum at 0.02 under Nettack with 2.0 perturbations), and then decreases as $\mu$ increases further. This trend suggests that classification accuracy initially improves and then deteriorates as the structural regularity of the training graphs declines. This phenomenon can be attributed to the fact that graphs with extremely high regularity (e.g., $\mu=0.00$) limit the diversity of neighborhood information, making models more prone to overfitting. Moderate irregularity introduces inter-community links that enhance generalization and robustness. In contrast, excessive irregularity (e.g., $\mu > 0.06$) introduces noise and disrupts the integrity of community structure, thereby hindering the model's ability to aggregate meaningful features.

\textbf{Guideline 1:} \textit{Training on highly regular graphs significantly undermines adversarial robustness, whereas introducing structural diversity in training graphs can enhance the model's robustness against adversarial attacks.}

\begin{table}[htbp]
	\centering
	\caption{Accuracy of GCN under adversarial training. AE Prop denotes the rate of adding adversarial examples.}
	\renewcommand{\arraystretch}{1.2}
	\resizebox{0.5\textwidth}{!}{ 
		\begin{tabular}{lllllll}
			\toprule
			\multirow{2}{*}{Dataset} & \multicolumn{6}{c}{AE Prop} \\ \cline{2-7} 
			& 0.00 & 0.02 & 0.04 & 0.06 & 0.08 & 0.10 \\ 
			\midrule
			Cora & 0.6865 & 0.6904 & \textbf{0.6913} & 0.6644 & 0.6558 & 0.6490 \\
			Citeseer & 0.5830 & 0.6250 & 0.6307 & \textbf{0.6409} & 0.6023 & 0.5932 \\
			PubMed & 0.7366 & 0.7500 & 0.7561 & \textbf{0.7726} & 0.7250 & 0.6726 \\
			Amazon Photo & 0.7056 & 0.7191 & \textbf{0.7225} & 0.6922 & 0.6856 & 0.6450 \\
			\bottomrule
		\end{tabular}
	}
	\label{table2_at}
\end{table}

\subsubsection{Adversarial Training and Pattern Diversity}
To counter adversarial attacks, a common approach is adversarial training, which augments the training data with generated adversarial examples and then retrains the model on the augmented data. According to \cite{wu2022ergcn}, the most effective way to perform adversarial attacks is to connect nodes with different features, which resemble irregular links in graphs. Hence, there is an internal consistency between adversarial training and the structural regularity of graphs. Table \ref{table2_at} presents the classification accuracy of GCN under adversarial training, where the proportion of adversarial examples added to the training data increases progressively. The results show that model performance improves as the proportion of adversarial examples increases, peaking at 0.04 or 0.06. Beyond this point, accuracy declines, suggesting that a moderate amount of adversarial examples improves robustness by increasing pattern diversity in training. However, excessive adversarial examples introduce noise, which reduces model stability and impairs generalization.

\textbf{Guideline 2:} \textit{The essence of adversarial training lies in increasing the pattern diversity within the training graphs of GNNs, which serves as an effective strategy for enhancing model adversarial robustness. }

\subsubsection{Graph Structural Characteristics of Adversarial Perturbation}
To gain a more comprehensive understanding of adversarial perturbations from the perspective of graph patterns, we systematically analyze their structural characteristics in GNNs by comparing perturbed components with their original components. Here, ten classic graph structure measures, including node degree, clustering coefficient, degree centrality, etc., are adopted, and their average values are calculated for each perturbed and unperturbed edge. As shown in Table \ref{table3_character}, adversarial perturbations generally exhibit higher structural measure values than the original components. Notably, betweenness, closeness, eigenvector, and edge betweenness centrality show significantly higher values under adversarial perturbations. This indicates a strong correlation with adversarial behavior and suggests that perturbations tend to exploit these centrality measures.

\begin{table}[ht]
	\setlength{\tabcolsep}{10pt}
	\centering
	\caption{Graph structural characteristics of adversarial attacks on GNNs (Mettack 10\% perturbation rate, Nettack 4.0 perturbations, and Random Attack 10\% perturbation rate). Note: $D$, degree; $C$, clustering coefficient; $DC$, degree centrality; $BC$, betweenness centrality; $CC$, closeness centrality; $EC$, eigenvector centrality; $KC$, Katz centrality; $ND$, neighbor degree; $EBC$, edge betweenness centrality; $ELC$, edge load centrality.}
	\renewcommand{\arraystretch}{1.1}
	\resizebox{1.0 \linewidth}{!}{
		\begin{tabular}{lllllll}
			\toprule
			\multirow{2}{*}{Operation} & \multicolumn{2}{c}{Mettack(0.10)} & \multicolumn{2}{c}{Nettack(4.0)} & \multicolumn{2}{c}{Random Attack(0.10)} \\ \cline{2-7} 
			& \begin{tabular}[c]{@{}l@{}}Clean\end{tabular} & \begin{tabular}[c]{@{}l@{}}Perturbation\end{tabular} & \begin{tabular}[c]{@{}l@{}}Clean\end{tabular} & \begin{tabular}[c]{@{}l@{}}Perturbation\end{tabular} & \begin{tabular}[c]{@{}l@{}}Clean\end{tabular} & \begin{tabular}[c]{@{}l@{}}Perturbation\end{tabular} \\ \midrule
			
			$D$ & {3.3580} & \textbf{5.5106} & {3.3984} & \textbf{4.9036} & {3.3488} & \textbf{4.5063} \\ 
			\begin{tabular}[l]{@{}l@{}}$C$\end{tabular} & {0.0056} & \textbf{0.0020} & {0.0039} & \textbf{0.0020} & {0.0058} & \textbf{0.0034} \\
			\begin{tabular}[l]{@{}l@{}}$DC$\end{tabular} & {0.0014} & \textbf{0.0022} & {0.0014}& \textbf{0.0020} & {0.0013} & \textbf{0.0018} \\ 
			\begin{tabular}[l]{@{}l@{}}$BC$\end{tabular} & {0.0015} & \textbf{0.0074} & {0.0025} & \textbf{0.0153} & {0.0017} & \textbf{0.0042} \\
			\begin{tabular}[l]{@{}l@{}}$CC$\end{tabular} & 0.1437 & \textbf{0.1679} & {0.1195} & \textbf{0.1397} & {0.1397} & \textbf{0.1485} \\ 
			\begin{tabular}[l]{@{}l@{}}$EC$\end{tabular} & {0.0085} & \textbf{0.0384} & {0.0063} & \textbf{0.0381} & {0.0144} & \textbf{0.0259} \\ 
			\begin{tabular}[l]{@{}l@{}}$KC$\end{tabular} & {0.0189} & \textbf{0.0252} & {0.0199} & \textbf{0.0240} & {0.0192} & \textbf{0.0219} \\ 
			\begin{tabular}[l]{@{}l@{}}$ND$\end{tabular} &{3.7119} & \textbf{4.8480} & {3.5351} & \textbf{4.2462} & {3.7521} & \textbf{3.9793} \\ 			
			\begin{tabular}[l]{@{}l@{}}$EBC$\end{tabular} & {0.0013} & \textbf{0.0037} & {0.0017} & \textbf{0.0083} & {0.0017} & \textbf{0.0036} \\ 
			\begin{tabular}[l]{@{}l@{}}$ELC$\end{tabular} & {12,049} & \textbf{33,532} & {12,225} & \textbf{26,854} & {13,520} & \textbf{23,793} \\ 
			\bottomrule
		\end{tabular}
	}
	\label{table3_character}
\end{table}

\textbf{Guideline 3:} \textit{Structural characteristics of graphs can serve as effective indicators for both the detection and mitigation of adversarial perturbations.}

\subsection{Effect of Model Architecture}
\subsubsection{Robustness Comparison of Model Architecture}
\begin{figure*}[htbp]
	\centering
	\includegraphics[width=\textwidth]{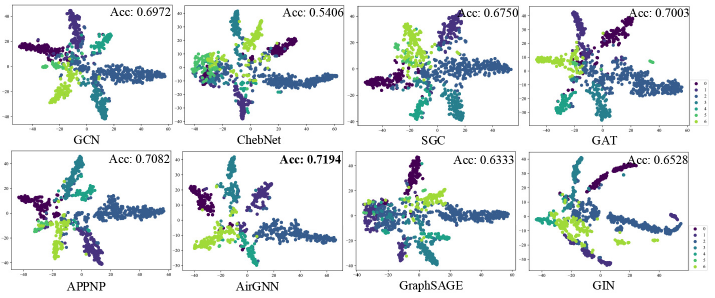}
	\caption{Accuracy of GNNs under Mettack with 5\% perturbation rate on Cora. The output spaces of the classification models are visualized using t-SNE. Colors represent different class labels; color separation shows the effectiveness of the models.}
	\label{fig4_model}
\end{figure*}

To systematically study the adversarial robustness of GNNs and provide a foundation for designing robust models, we evaluate representative architectures across five categories, i.e., Convolutional-GNNs, Attention-GNNs, Diffusion-GNNs, Sampling-GNNs, and Isomorphism-GNNs, under typical adversarial attack settings. The t-SNE visualizations of the decision space for these models are shown in Fig. \ref{fig4_model}. Each subfigure presents the test outputs, including node embeddings, class labels, and accuracy under Mettack (5\% perturbation rate). Node colors represent class labels, and the best classification accuracy is highlighted in bold. From the visualizations and quantitative performance variations, we observe that under identical adversarial attack, the AirGNN model exhibits the clearest decision boundaries and achieves the highest classification accuracy, followed by APPNP. In contrast, models such as ChebNet, GraphSAGE, and GIN exhibit less distinct decision boundaries and lower accuracy. 

\begin{table*}[htbp]
	\centering
	\setlength{\tabcolsep}{10pt}
	\caption{Accuracy of GNNs under Mettack on four classical datasets. \textbf{Bold} numbers show the best results of all methods. }
	\renewcommand{\arraystretch}{1.1}  
	\resizebox{\textwidth}{!}{ 
		\begin{tabular}{llllllllll}
			\hline
			\multirow{2}{*}{Dataset} & \multirow{2}{*}{Ptb Rate} & \multicolumn{8}{c}{Model} \\ \cline{3-10} 
			& & GCN & ChebNet & SGC & GAT & APPNP & AirGNN & GraphSAGE & GIN \\ \hline
			\multirow{6}{*}{Cora} & 0.00 & 0.7352 & 0.5925 & 0.7230 & 0.7551 & 0.7623 & \textbf{0.7655} & 0.6621 & 0.7021 \\
			& 0.02 & 0.7223 & 0.5804 & 0.7149 & 0.7238 & 0.7430 & \textbf{0.7536} & 0.6325 & 0.6843 \\
			& 0.04 & 0.7113 & 0.5520 & 0.6904 & 0.7028 & 0.7168 & \textbf{0.7272} & 0.6306 & 0.6731 \\
			& 0.06 & 0.6891 & 0.5514 & 0.6609 & 0.6905 & 0.6937 & \textbf{0.7043} & 0.6271 & 0.6490 \\
			& 0.08 & 0.6718 & 0.5356 & 0.6213 & 0.6735 & 0.6819 & \textbf{0.6916} & 0.6159 & 0.6362 \\
			& 0.10 & 0.6455 & 0.5267 & 0.6019 & 0.6471 & 0.6599 & \textbf{0.6764} & 0.6029 & 0.6052 \\ \hline
			\multirow{6}{*}{Citeseer} & 0.00 & 0.6710 & 0.5300 & 0.6702 & 0.6724 & 0.6845 & \textbf{0.6974} & 0.6474 & 0.6373 \\
			& 0.02 & 0.6525 & 0.5290 & 0.6562 & 0.6610 & 0.6684 & \textbf{0.6834} & 0.6194 & 0.6045 \\
			& 0.04 & 0.6198 & 0.5202 & 0.6151 & 0.6451 & 0.6485 & \textbf{0.6682} & 0.6165 & 0.5715 \\
			& 0.06 & 0.5965 & 0.5172 & 0.5850 & 0.6297 & 0.6360 & \textbf{0.6535} & 0.5927 & 0.5613 \\
			& 0.08 & 0.5747 & 0.5140 & 0.5740 & 0.6042 & 0.6108 & \textbf{0.6315} & 0.5865 & 0.5452 \\
			& 0.10 & 0.5541 & 0.5012 & 0.5516 & 0.5882 & 0.5887 & \textbf{0.6126} & 0.5615 & 0.5053 \\ \hline
			\multirow{6}{*}{PubMed} & 0.00 & 0.7823 & 0.7396 & 0.7911 & 0.7841 & 0.7929 & \textbf{0.7963} & 0.7765 & 0.7733 \\
			& 0.02 & 0.7743 & 0.7376 & \textbf{0.8044} & 0.7796 & 0.7916 & 0.7932 & 0.7749 & 0.7690 \\
			& 0.04 & 0.7664 & 0.7109 & \textbf{0.8040} & 0.7627 & 0.7863 & 0.7843 & 0.7582 & 0.7660 \\
			& 0.06 & 0.7621 & 0.6843 & \textbf{0.8036} & 0.7529 & 0.7812 & 0.7794 & 0.7462 & 0.7659 \\
			& 0.08 & 0.7513 & 0.6638 & \textbf{0.8031} & 0.7430 & 0.7770 & 0.7731 & 0.7315 & 0.7616 \\
			& 0.10 & 0.7329 & 0.6099 & \textbf{0.8029} & 0.7345 & 0.7654 & 0.7637 & 0.6886 & 0.7441 \\ \hline
			\multirow{6}{*}{\begin{tabular}[l]{@{}c@{}}Amazon\\ Photo\end{tabular}} & 0.00 & 0.8697 & 0.4916 & 0.9164 & 0.9169 & \textbf{0.9208} & 0.8752 & 0.6508 & 0.2957 \\
			& 0.02 & 0.8332 & 0.4593 & 0.8522 & 0.8696 & \textbf{0.9002} & 0.8409 & 0.6073 & 0.2948 \\
			& 0.04 & 0.7677 & 0.4212 & 0.7998 & 0.8600 & \textbf{0.8810} & 0.8290 & 0.5861 & 0.2856 \\
			& 0.06 & 0.6832 & 0.3765 & 0.7642 & 0.8322 & \textbf{0.8674} & 0.7830 & 0.5121 & 0.2806 \\
			& 0.08 & 0.6352 & 0.3478 & 0.7316 & 0.7695 & \textbf{0.8205} & 0.7399 & 0.5013 & 0.2776 \\
			& 0.10 & 0.6090 & 0.3385 & 0.6906 & 0.7052 & \textbf{0.8016} & 0.6959 & 0.4880 & 0.2739 \\ \hline
		\end{tabular}
	}
	\label{table4_acc}
\end{table*}

\begin{table*}[htbp]
	\centering
	\caption{Accuracy of GNNs under classical adversarial attacks. \textbf{Bold} numbers show the best results of all methods. }
	\setlength{\tabcolsep}{10pt}
	\renewcommand{\arraystretch}{1.2}  
	\resizebox{\textwidth}{!}{ 
		\begin{tabular}{llllllllll}
			\midrule
			\multirow{2}{*}{Method(Ptb Rate/Num)} & \multirow{2}{*}{Dataset} & \multicolumn{8}{c}{Model} \\ \cline{3-10} 
			& & GCN & ChebNet & SGC & GAT & APPNP & AirGNN & GraphSAGE & GIN \\ 
			\midrule
			\multirow{3}{*}{Mettack(0.05)} & Cora & 0.6972 & 0.5406 & 0.6750 & 0.7003 & 0.7082 & \textbf{0.7194} & 0.6333 & 0.6528 \\
			& Citeseer & 0.6174 & 0.5224 & 0.6022 & 0.6357 & 0.6436 & \textbf{0.6619} & 0.5961 & 0.5792 \\
			& PubMed & 0.7660 & 0.6956 & \textbf{0.8036} & 0.7679 & 0.7821 & 0.7852 & 0.7504 & 0.7650 \\ 
			& Amazon Photo & 0.7405 & 0.4005 & 0.7829 & 0.8588 & \textbf{0.8743} & 0.8163 & 0.5518 & 0.2828 \\ 
			\midrule
			\multirow{3}{*}{Nettack(2.0)} & Cora & 0.6562 & 0.5813 & 0.6813 & \textbf{0.8156} & 0.7104 & 0.7031 & 0.5912 & 0.5844 \\
			& Citeseer & 0.6536 & 0.5522 & 0.6376 & \textbf{0.8571} & 0.6640 & 0.7786 & 0.5852 & 0.6214 \\
			& PubMed & 0.2826 & 0.7380 & 0.7463 & \textbf{0.7783} & 0.7779 & 0.3000 & 0.7690 & 0.2870 \\ 
			& Amazon Photo & 0.7250 & 0.4290 & 0.9120 & 0.8925 & \textbf{0.9154} & 0.7575 & 0.6232 & 0.4290 \\ 
			\midrule
			\multirow{3}{*}{Random Attack(0.05)} & Cora & 0.7236 & 0.5902 & 0.7077 & 0.7302 & 0.7320 & \textbf{0.7350} & 0.7226 & 0.6764 \\
			& Citeseer & 0.6548 & 0.5184 & 0.6390 & 0.6561 & 0.6568 & \textbf{0.6583} & 0.6545 & 0.6160 \\
			& PubMed & 0.7759 & 0.7378 & 0.7446 & 0.7799 & 0.7882 & \textbf{0.7895} & 0.7749 & 0.7668 \\ 
			& Amazon Photo & 0.8041 & 0.4607 & 0.8985 & 0.8964 & \textbf{0.9007} & 0.8175 & 0.6930 & 0.2971 \\ 
			\bottomrule
		\end{tabular}
	}
	\label{table5_acc}
\end{table*}

These trends are supported by the results in Tables \ref{table4_acc} and \ref{table5_acc}, providing a comprehensive evaluation of adversarial robustness across multiple datasets and attack scenarios. Table \ref{table4_acc} shows that classification accuracy generally declines as the perturbation rate increases. On Cora and Citeseer, Diffusion-GNNs (AirGNN and APPNP) consistently perform well across perturbation levels, with APPNP achieving the highest accuracy on Amazon Photo, confirming the robustness advantage of Diffusion-GNNs. In contrast, ChebNet exhibits the weakest adversarial robustness. SGC performs best on PubMed, suggesting that its resilience aligns well with the structural properties of the dataset. Table \ref{table5_acc} reinforces these observations under fixed attack configurations. Under Mettack with a 5\% perturbation rate, AirGNN demonstrates strong robustness on Cora and Citeseer, while SGC achieves the highest accuracy on PubMed. On Amazon Photo, APPNP attains the best result, highlighting Diffusion-GNNs' strength in capturing global structural information. Under Nettack with a perturbation budget of 2.0, GAT leads on Cora, Citeseer, and PubMed, while APPNP performs best on Amazon Photo. However, AirGNN shows a noticeable performance drop on PubMed under Nettack, indicating a potential vulnerability to structure-focused perturbations. Under the Random Attack (5\% perturbation rate), both AirGNN and APPNP maintain high accuracy, especially on Amazon Photo.

The variations in adversarial robustness among model architectures can be attributed to their distinct designs and message-passing mechanisms. Diffusion-GNNs, such as APPNP and AirGNN, propagate information iteratively across the graph and employ teleportation mechanisms or residual mixing. This enables them to preserve global structural information better and reduce their dependence on individual edges or neighbors, thus mitigating the impact of localized perturbations and improving robustness. In contrast, models like ChebNet, GraphSAGE, and GIN primarily rely on local aggregation, making them more vulnerable to adversarial perturbations targeting neighborhood structures. Attention-GNNs, which dynamically assign weights to neighbors, often perform well, especially as source models in transfer-based attacks. However, their attention mechanisms may be sensitive to adversarial changes, particularly when perturbed features disproportionately influence attention scores, leading to degraded performance.

\textbf{Guideline 4:} \textit{Diffusion-GNNs, such as APPNP and AirGNN, show strong resistance to adversarial perturbations, making diffusion mechanisms a promising choice for improving adversarial robustness.} 

\textbf{Guideline 5:} \textit{Sampling-GNNs, such as GraphSAGE, and Isomorphism-GNNs, such as GIN, are highly vulnerable to adversarial perturbations, while Convolutional-GNNs, including GCN, ChebNet, and SGC, exhibit inconsistent adversarial robustness. These architectures should be carefully evaluated when designing robust GNNs} 

\begin{figure*}[hp]
	\centering
	\includegraphics[width=\textwidth]{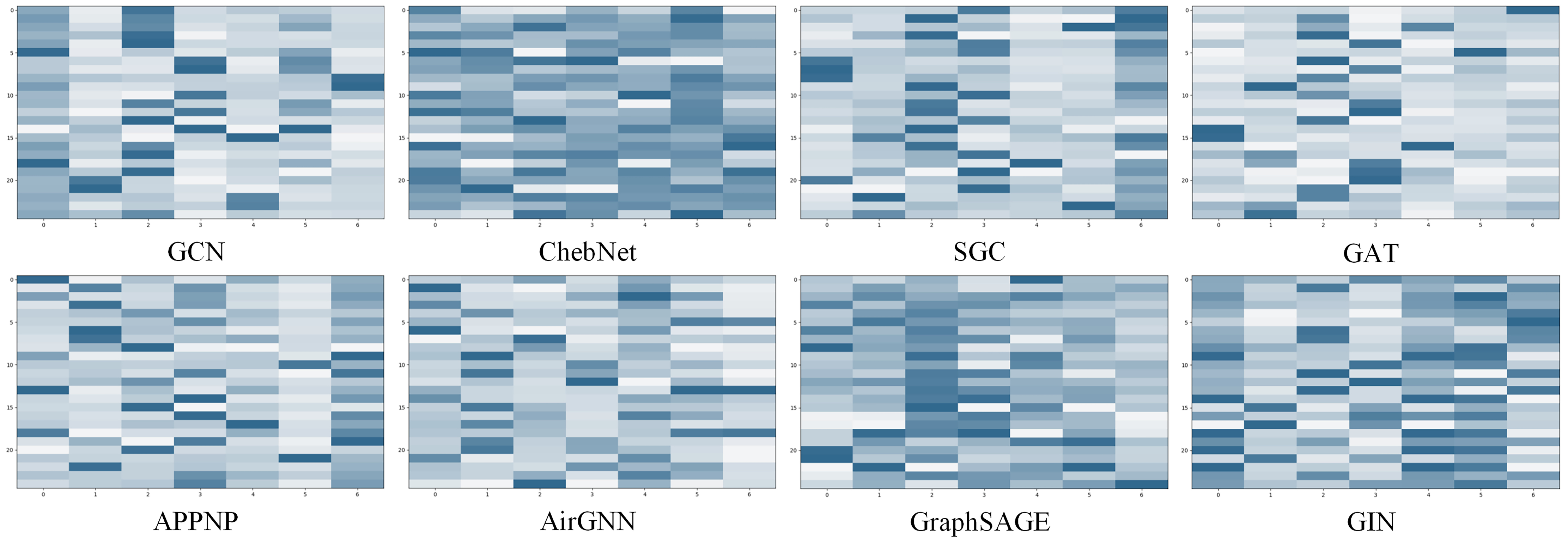}
	\caption{Heatmap visualization of GNNs' layer-2 node representations on 25 perturbed nodes under 5\% Mettack. Rows in each subfigure represent perturbed nodes, and columns represent feature dimensions from the second (hidden) layer. Darker colors indicate higher activation values. More consistent and smoother activation distributions indicate stronger adversarial robustness.}	
	\label{fig5_heatmap}
\end{figure*}

\subsubsection{Visualization of Hidden Representations}
To further investigate the internal mechanisms underlying the adversarial robustness of different GNNs, we visualize the hidden node representations under adversarial attacks using heatmaps (Fig. \ref{fig5_heatmap}). Specifically, we visualize the second-layer hidden representations of each GNN for 25 perturbed nodes under a 5\% Mettack perturbation rate. The rows in each subfigure represent nodes, and the columns correspond to feature dimensions. Darker colors indicate higher activation values.
These visualizations reveal notable differences across models. Diffusion-GNNs, such as AirGNN and APPNP, display smoother and more consistent activation distributions, indicating their hidden representations remain more stable under adversarial perturbations. This stability aligns with their diffusion-based architecture, which facilitates consistent information propagation across layers and supports iterative feature refinement through repeated context integration. In contrast, Sampling-GNNs (e.g., GraphSAGE) and Isomorphism-GNNs (e.g., GIN) rely on local aggregation mechanisms that emphasize neighborhood features, limiting their ability to integrate global structural information. This reduces their robustness to adversarial perturbations, making their hidden representations more susceptible to perturbations. 

\textbf{Guideline 6: } \textit{The stability of hidden representations under adversarial perturbations is closely related to adversarial robustness. Maintaining smooth node activation distributions is indicative of stronger adversarial robustness.} 

\subsubsection{Analysis of Decision Surface}
\begin{figure}[htbp]
	\centering
	\includegraphics[width=\columnwidth]{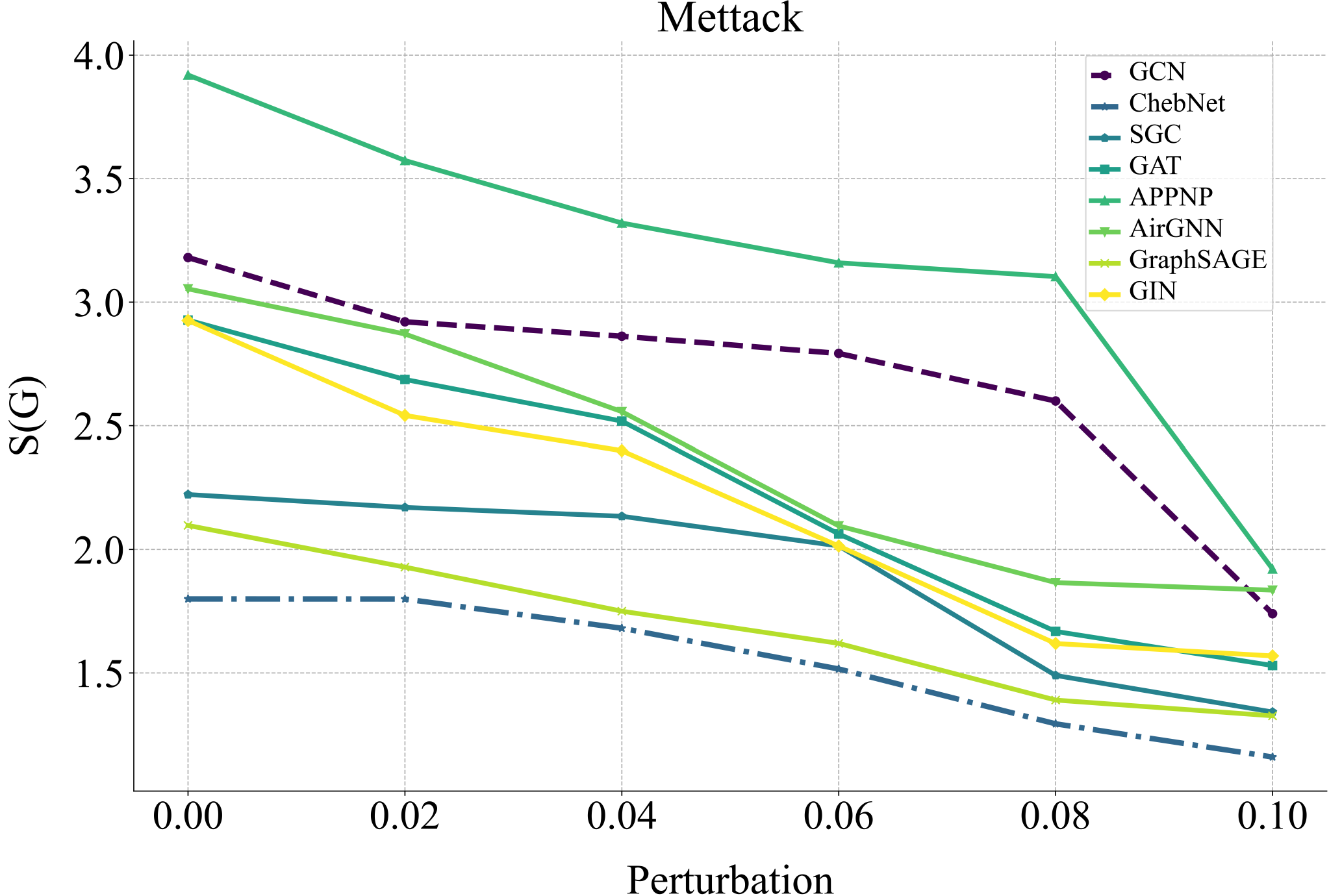}
	\caption{${\mathcal S}\left( {\mathcal G} \right)$ of GNNs under Mettack on Cora.}
	\label{fig6_db}
\end{figure}

In order to analyze the adversarial robustness of GNNs from the perspective of decision boundary, the ``decision surface'' ${\mathcal S}\left( {\mathcal G} \right)$ is used as a metric to quantify the difference between predictions for different node classes. Adversarial attacks fool target models by using perturbations that push input samples across the decision boundary. Therefore, a robust model should have an explicit decision boundary and maintain a high and stable ${\mathcal S}\left( {\mathcal G} \right)$ value as attack intensity increases. The values of ${\mathcal S}\left( {\mathcal G} \right)$ for GNNs under Mettack on Cora are shown in Fig. \ref{fig6_db}. According to the results, the APPNP model has the highest ${\mathcal S}\left( {\mathcal G} \right)$ value, indicating the strongest adversarial robustness. The lower ${\mathcal S}\left( {\mathcal G} \right)$ values of ChebNet and GraphSAGE imply weaker adversarial robustness. Moreover, the ${\mathcal S}\left( {\mathcal G} \right)$ value of GCN exhibits instability. This finding aligns with previous experimental results.

\textbf{Guideline 7:} \textit{Robust GNNs tend to exhibit clear decision boundaries. Enhancing the separability between classes contributes significantly to adversarial robustness.}

\subsubsection{Neuron-Level Analysis of GNNs}
To understand adversarial attack behavior in GNNs from the perspective of neural activation, we analyze changes in neuron activation before and after attacks to identify the sensitive neurons that contribute significantly to incorrect predictions. If hidden-layer neurons remain stable, exhibiting only minor changes in activation under adversarial perturbations, the model is likely to produce robust representations and accurate predictions. Therefore, if sensitive neurons constitute only a small fraction of the total, adversarial robustness can be enhanced by locating and repairing them. To identify sensitive neurons, given a robust neuron function ${{\mathcal M}}( \cdot )$, if two samples ${x_1}$ and ${x_2}$ from the dataset ${\mathcal D}$ are similar, they are expected to produce similar outputs \cite{zhang2020interpreting}.
\begin{equation}
	\text{if } \|{x_1} - {x_2}\| \le \varepsilon \Rightarrow \|{{\mathcal M}}({x_1}) - {{\mathcal M}}({x_2})\| \le \delta
\end{equation}
where $\| \cdot \|$ denotes a distance metric to quantify the distance between sample pairs, and $\varepsilon$ and $\delta$ are small values. 

Due to the imperceptibility of adversarial perturbations, the benign sample ${x_i}$ and its corresponding adversarial example ${{\hat x}_i}$ should be similar and expected to satisfy this condition. However, in practice, ${x_i}$ and ${{\hat x}_i}$ often produce significantly different representations due to the non-robustness of neurons. Thus, neuron sensitivity can be quantified by the deviation between benign and adversarial representations:
\begin{equation}
	\sigma ({{\mathcal M}},\bar {\mathcal D}) = \frac{1}{N}\sum\limits_{i = 1}^N {\frac{1}{{\mathrm{dim} ({{\mathcal M}}({x_i}))}}\|{{\mathcal M}}({x_i}) - {{\mathcal M}}({{\bar x}_i})\|}
\end{equation}
where $\bar{\mathcal D} = \{ ({x_i},{{\bar x}_i}) \mid i=1,2,\ldots,N \}$ denotes the set of sample pairs, and $\mathrm{dim}  (\cdot)$ denotes the vector dimension. Consequently, larger $\sigma $ indicate higher neuron sensitivity and can serve as an effective basis for adversarial analysis.

\begin{figure}[htbp]
	\centering
	\includegraphics[width=\columnwidth]{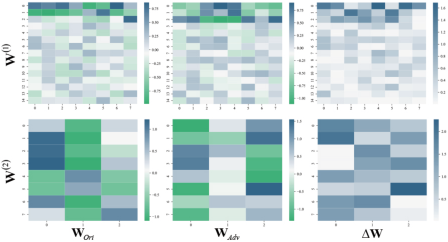}
	\caption{The weights ${\mathbf{W}_{Ori}}$ and ${\mathbf{W}_{Adv}}$ of the two-layer GCN before and after the adversarial attack, along with the weight difference $\Delta \mathbf{W}$. ${\mathbf{W}^{(1)}}$ and ${\mathbf{W}^{(2)}}$ denote the weight visualizations of the first and second hidden layers of GCN, respectively.} 
	\label{fig7_weight}
\end{figure}

We explore the adversarial robustness of GNNs from the perspective of neuron sensitivity and measure the extent to which neurons change under benign and adversarial examples. Since the weight matrix ${\mathbf{W}}$ of GNNs reflects the learning process and state changes of neurons, in this study, we quantify neuron sensitivity by analyzing changes in ${\mathbf{W}}$. Specifically, for a two-layer GCN, Fig. \ref{fig7_weight} shows the weights before and after the adversarial attack, namely ${\mathbf{W}_{Ori}}$, ${\mathbf{W}_{Adv}}$, and the weight difference $\Delta \mathbf{W}$. The first and second rows represent the weights of the first and second hidden layers, ${\mathbf{W}^{(1)}}$ and ${\mathbf{W}^{(2)}}$, respectively. We assume that the more robust the target model is, the more consistent its weight matrices remain under adversarial attacks. In other words, for a robust model, the $\mathbf{W}$ displayed in the first and second columns of Fig. \ref{fig7_weight} should be similar. The third column in Fig. \ref{fig7_weight}, represented by $\Delta \mathbf{W}$, visualizes the changes in model weights before and after adversarial attacks, where darker regions indicate greater weight changes. The results indicate that adversarial perturbations mainly affect a small subset of weights, while most remain unchanged.

\textbf{Guideline 8:} \textit{Only a small portion of neurons in GNNs are affected by adversarial perturbations, while most remain stable. Therefore, locating and repairing sensitive neurons can enhance robustness against adversarial attacks. } 

\subsubsection{Transferability of Adversarial Examples}
Recent studies have shown that adversarial examples exhibit cross-model transferability and can effectively attack target models in black-box scenarios with a high success rate. According to these findings, adversarial examples generated from a specific model can fool other models with different architectures and training data. This property of adversarial examples has inspired the development of black-box adversarial attacks, where a substitute model (source model) is trained to simulate the target model and then used to generate adversarial examples without interacting with the target model. Hence, understanding the essence of adversarial transferability is an important problem for improving the adversarial robustness of GNNs. 

\begin{figure}[htbp]
	\centering
	\includegraphics[width=\columnwidth]{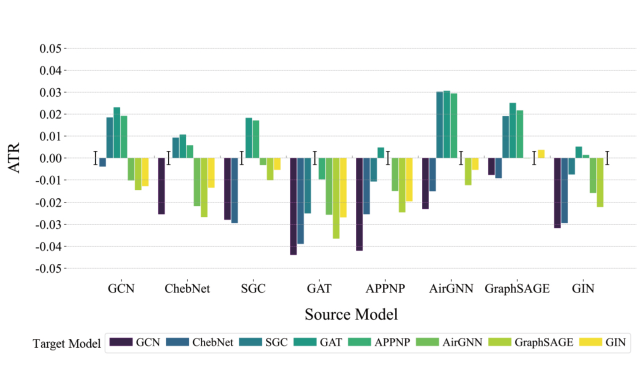}
	\caption{Statistical results of ATR under transfer-based attacks with different model architectures. The horizontal axis denotes the source model, the vertical axis denotes the ATR, and the legend denotes the target model.
	} 
	\label{fig8_atr_model}
\end{figure}
\begin{figure}[h]
	\centering
	\begin{subfigure}[b]{\columnwidth}
		\centering
		\includegraphics[width=\columnwidth]{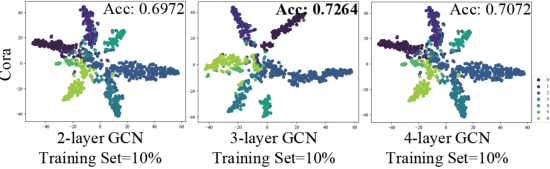}
		\caption{}
		\label{fig9_capacity_a}
	\end{subfigure}
	\begin{subfigure}[b]{\columnwidth}
		\centering
		\includegraphics[width=\columnwidth]{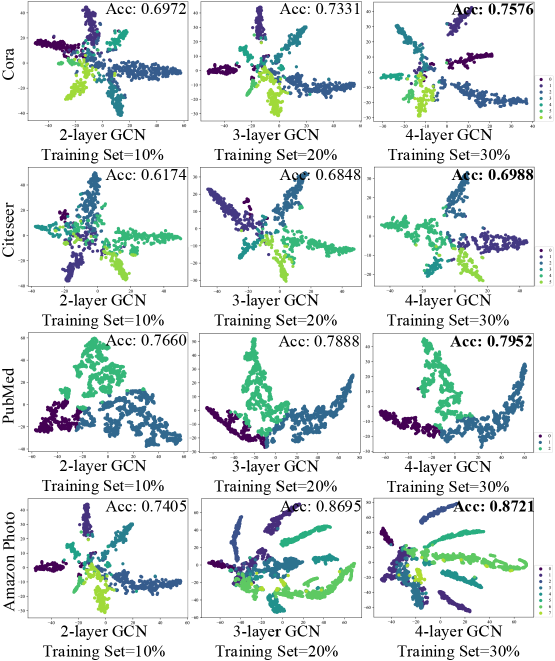}
		\caption{}
		\label{fig9_capacity_b}
	\end{subfigure}
	\caption{Accuracy of two-, three-, and four-layer GCN under Mettack (5\% perturbation rate). (a) Adversarial robustness evaluation on Cora with a fixed 10\% training data. (b) Adversarial robustness across Cora, Citeseer, PubMed, and Amazon Photo with training data scaled alongside model depth.}
	\label{fig9_capacity}
\end{figure}
\begin{figure}[htbp]
	\centering
	\includegraphics[width=0.5\columnwidth]{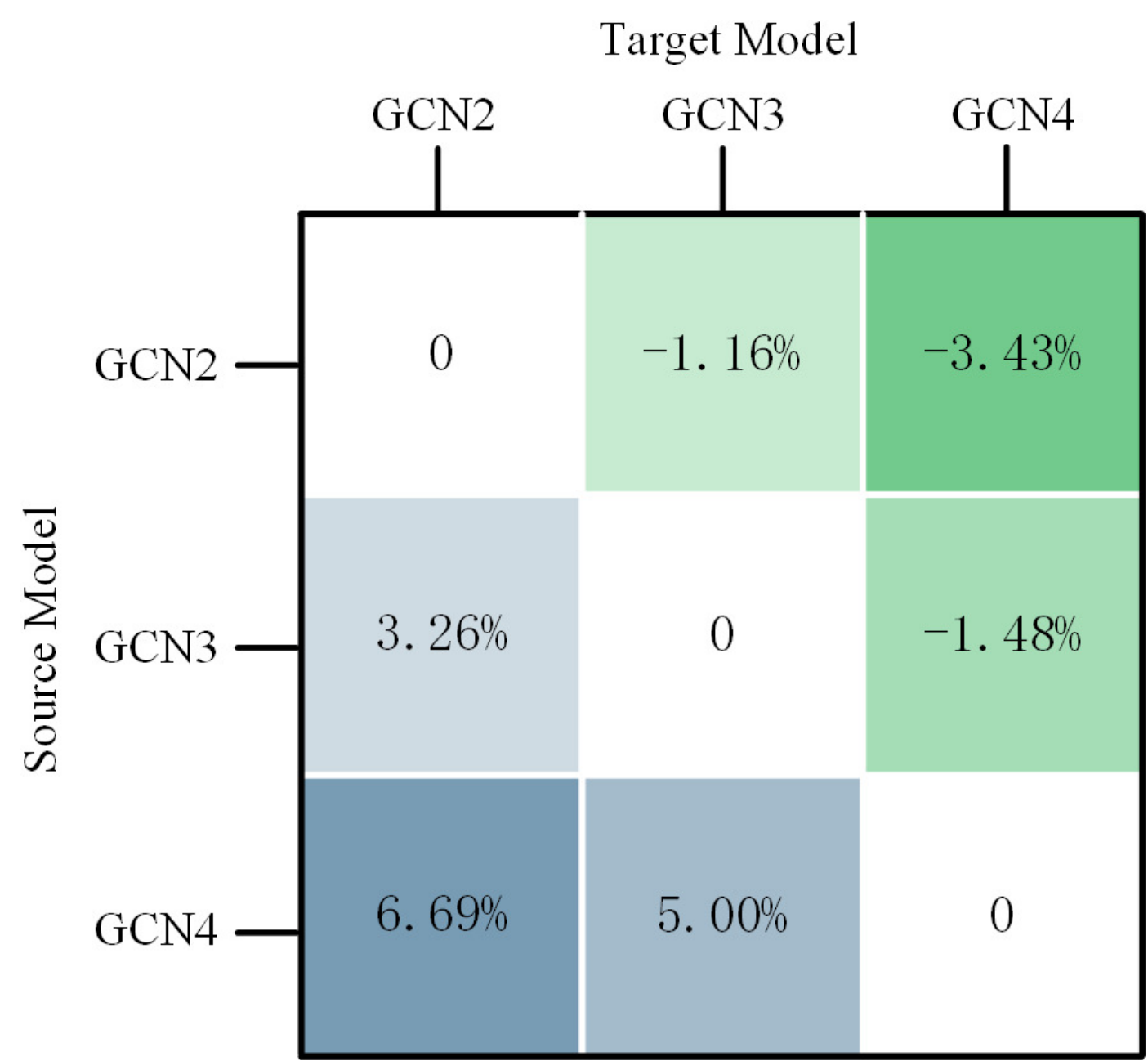}
	\caption{Transferability of adversarial examples under different model capacities. The smaller the value of ATR, the more transferable the adversarial examples are. } 
	\label{fig10_atr_capacity}
\end{figure}

To investigate how model architecture affects the transferability of adversarial examples, we conduct empirical experiments by selecting a set of classical GNNs and performing transfer-based attacks among them. Adversarial examples are generated by first attacking the source model and then transferring these examples to fool the target models. The ATR values based on the accuracy of the models under Mettack with a 5\% perturbation rate are shown in Fig. \ref{fig8_atr_model}. According to Equation \eqref{eq_atr} in Section \ref{sec_Evaluation Metrics}, smaller ATR values indicate greater transferability. Fig. \ref{fig8_atr_model} shows that when GAT is used as the source model, adversarial examples transferred to other models result in the lowest ATR values, indicating the highest transferability. When GCN is the target model, it shows weak robustness, as reflected by its low ATR values, which means it is easily affected by adversarial examples from other models. Moreover, except for GAT, adversarial examples generated from other source models generally exhibit weak transferability. In particular, when GAT and APPNP serve as target models, transfer-based attacks are the least effective, further demonstrating their strong adversarial robustness.

\textbf{Guideline 9:} \textit{Adversarial examples from Attention-GNNs transfer more effectively to other models, while both Attention- and Diffusion-GNNs show stronger robustness as targets.}

\subsection{Effect of Model Capacity}
\subsubsection{Robustness Comparison of Model Capacity}
For the purpose of investigating the relationship between model capacity and the adversarial robustness of GNNs, we systematically compare the classification accuracy of two-, three-, and four-layer GCNs under three representative adversarial attack methods. 

The accuracy on Cora under Mettack with a 5\% perturbation rate is shown in Fig. \ref{fig9_capacity_a}. The results indicate that increasing model capacity initially improves classification accuracy, which then declines. This suggests that there is no straightforward correlation between model capacity and adversarial robustness. We believe that model complexity increases with model capacity, and learning the decision boundaries of high-capacity models requires a larger number of training samples. Thus, training models of different capacities with the same training data is unreasonable. It is essential to scale the amount of training data according to model capacity. Fig. \ref{fig9_capacity_b} shows the classification accuracy of two-, three-, and four-layer GCNs on four datasets under Mettack, with training data increasing simultaneously. The results indicate that as the scale of training data increases, the classification accuracy improves with the number of hidden layers. Additionally, a comparison of GCN results on Cora in Fig. \ref{fig9_capacity} reveals that, for models of the same capacity, increasing the training data leads to clearer decision boundaries and improved accuracy. This confirms that higher-capacity models benefit from larger training data. Higher-capacity models offer greater expressive power but are more prone to overfitting or instability when trained on limited data. Thus, it is crucial to balance model capacity and the amount of available training data to ensure robust and stable performance. 

\textbf{Guideline 10:} \textit{Models with higher capacity tend to be more robust against adversarial attacks when sufficient training data is provided. Thus, increasing model capacity is an effective strategy to improve adversarial robustness in data-rich settings.}

\subsubsection{Transferability of Adversarial Examples}
In order to analyze the impact of model capacity on transfer-based attacks, Fig. \ref{fig10_atr_capacity} presents the ATR values when GCNs with different capacities (i.e., two-, three-, and four-layer GCNs) are used as source and target models, respectively. The results show that the transferability of adversarial examples exhibits asymmetry. Specifically, the values in the upper triangle of Fig. \ref{fig10_atr_capacity} are negative, whereas those in the lower triangle are positive. Moreover, in terms of model capacity, the results demonstrate that a transfer-based adversarial attack using GCN2 as the source model produces lower ATR values compared to those generated using GCN3 and GCN4. Conversely, as shown in the third column of Fig. \ref{fig10_atr_capacity}, attacks targeting GCN4 result in lower ATR values than those targeting GCN2 and GCN3. 

\textbf{Guideline 11:} \textit{Transfer-based adversarial attacks tend to be more effective when the source model has a smaller capacity and the target model has a larger capacity. To improve adversarial robustness in such scenarios, GNNs should avoid deploying overly large models as target models.}

\section{Conclusions}
\label{sec6}
In this study, we comprehensively explored the robustness of GNNs against adversarial attacks, using node classification as the target task. Through extensive experiments on mainstream GNNs, we systematically examined how graph patterns, model architectures, and model capacity affect adversarial robustness. Additionally, we proposed two novel metrics to quantitatively analyze neuron sensitivity and the transferability of adversarial examples. This thorough analysis enabled us to derive eleven actionable design guidelines for constructing robust GNNs, providing valuable insights for future research and practical applications. In future work, it is worth exploring the adversarial robustness of GNNs under more diverse attack settings, such as evasion attacks and backdoor attacks. Moreover, while classification accuracy under adversarial attacks is used to evaluate robustness indirectly, there remains a need for evaluation metrics that assess the robustness of GNNs independently of specific attacks.

\printcredits
\bibliographystyle{elsarticle-num-names}

\bibliography{GNNs-Robustness-Exploration}

\end{document}